\pdfoutput=1

\documentclass[11pt]{article}

\usepackage[]{acl}

\usepackage{times}
\usepackage{latexsym}

\usepackage[T1]{fontenc}

\usepackage[utf8]{inputenc}
\usepackage{microtype}

\usepackage{graphicx}
\usepackage{multirow}
\usepackage{mathtools}
\usepackage{diagbox}
\usepackage{float}
\usepackage{color}
\usepackage[normalem]{ulem}
\usepackage{amsfonts}
\usepackage{amsmath}
\usepackage{nccmath}
\usepackage[linesnumbered, ruled, noend]{algorithm2e}
\usepackage[noend]{algpseudocode}
\usepackage{stfloats}

\newcommand\name{HLP}
\newcommand\EVI{Evidence Existence}
\newcommand\evi{evidence existence}

\title{Hyperlink-induced Pre-training \\ for Passage Retrieval in Open-domain Question Answering}

\author{Jiawei Zhou$^1$, Xiaoguang Li$^2$, Lifeng Shang$^2$, Lan Luo$^3$, Ke Zhan$^3$, Enrui Hu$^3$, 
\\
{\bf Xinyu Zhang$^3$, Hao Jiang$^3$, Zhao Cao$^3$, Fan Yu$^3$, Xin Jiang$^2$, Qun Liu$^2$, Lei Chen$^1$} 
\\
$^1$The Hong Kong University of Science and Technology \\
$^2$Huawei Noah’s Ark Lab \qquad $^3$Distributed and Parallel Software Lab, Huawei
\\
\texttt{\small $^1$\{jzhoubu,leichen\}@ust.hk} 
\\
\texttt{\small $^{2,3}$\{lixiaoguang11,Shang.Lifeng,luolan13,zhanke2,huenrui1,zhangxinyu35} \\ \texttt{\small jianghao66,caozhao1,fan.yu,Jiang.Xin,qun.liu\}@huawei.com}
}

\begin{document}
\maketitle
\begin{abstract}

To alleviate the data scarcity problem in training question answering systems, recent works propose additional intermediate pre-training for dense passage retrieval~(DPR). However, there still remains a large discrepancy between the provided upstream signals and the downstream question-passage relevance, which leads to less improvement. To bridge this gap, we propose the \textbf{H}yper\textbf{L}ink-induced \textbf{P}re-training (\name), a method to pre-train the dense retriever with the text relevance induced by hyperlink-based topology within Web documents. We demonstrate that the hyperlink-based structures of \textit{dual-link} and \textit{co-mention} can provide effective relevance signals for large-scale pre-training that better facilitate downstream passage retrieval. We investigate the effectiveness of our approach across a wide range of open-domain QA datasets under zero-shot, few-shot, multi-hop, and out-of-domain scenarios. The experiments show our \name~outperforms the BM25 by up to 7 points as well as other pre-training methods by more than 10 points in terms of top-20 retrieval accuracy under the zero-shot scenario. Furthermore, \name~significantly outperforms other pre-training methods under the other scenarios. 
~
\let\thefootnote\relax\footnote{Our code and trained models are available at \url{https://github.com/jzhoubu/HLP}.}
~
\begin{figure}[tb]
	\begin{center}
		\includegraphics[width=0.48\textwidth]{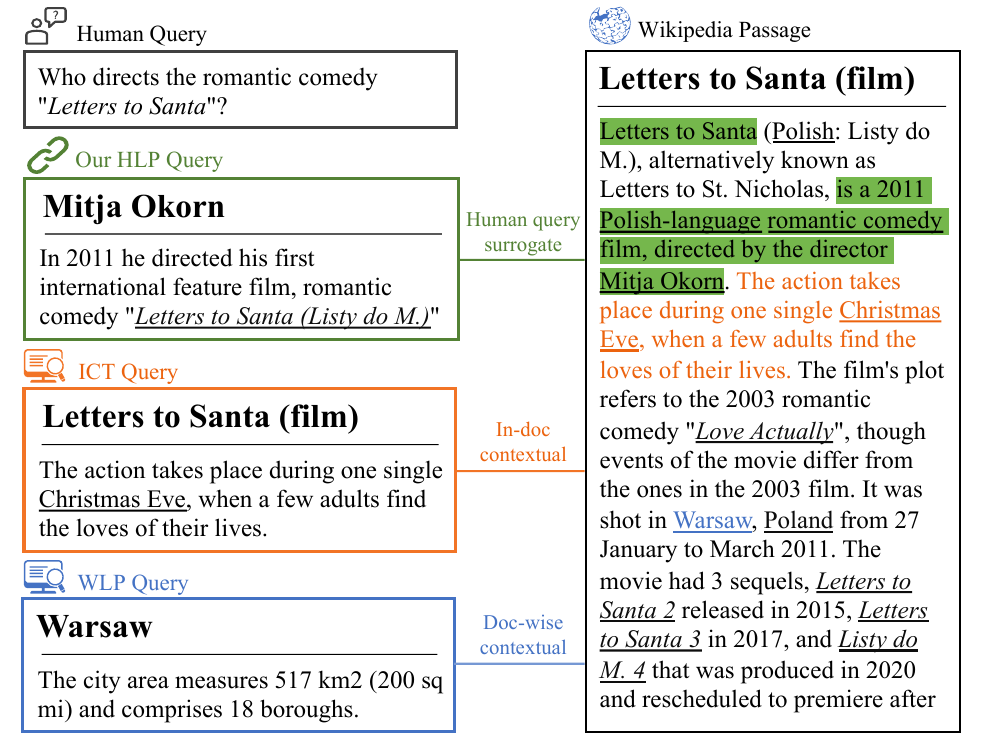}
		\caption{An example of different kinds of pseudo Q-P pairs. \underline{Underlined} texts are hypertexts that linked to other Wikipedia pages. The ICT query is a random sentence originated from the passage and the WLP query is a sentence from the first section of an out-link document of the given passage. The text highlighted in green gives evidence to answer the human query, and our proposed \name~query can be a better surrogate of the human query.}
		\label{Fig:img1}
	\end{center}
	\vspace{-0.8cm}
\end{figure}

\end{abstract}

\section{Introduction}
\label{sec:intro}
Open-domain question answering (OpenQA) aims to answer factual open questions with a large external corpus of passages. Current approaches to OpenQA usually adopt a two-stage retriever-reader paradigm~\cite{chen2017reading, zhu2021retrieving} to fetch the final answer span. The performance of OpenQA systems is largely bounded by the retriever as it determines the evidential documents for the reader to examine. Traditional retrievers, such as TF-IDF and BM25 \cite{robertson2009probabilistic}, are considered incapable of adapting to scenarios where deep semantic understanding is required. Recent works \cite{lee2019latent, karpukhin2020dense, qu2021rocketqa} show that by fine-tuning pre-trained language models on sufficient downstream data, dense retrievers can significantly outperform traditional term-based retrievers. 

Considering the data-hungry nature of the neural retrieval models, extensive efforts~ \cite{lee2019latent,chang2020pre,sachan2021end} have been made to design self-supervised tasks to pre-train the retriever. However, these pre-training tasks construct relevance signals largely depending on easily attainable sentence-level or document-level contextual relationships. For example, the relationship between a sentence and its originated context~(shown by the ICT query in Figure \ref{Fig:img1}) may not be sufficient enough to facilitate question-passage matching for the tasks of OpenQA. We also find that these pre-trained retrievers still fall far behind BM25 in our pilot study on the zero-shot experiment.

In order to address the shortcomings of the matching-oriented pre-training tasks as mentioned above, we propose a pre-training method with better surrogates of real natural question-passage (Q-P) pairs. We consider two conditions of relevance within Q-P pairs, which is similar to the process of distantly supervised retriever learning~\cite{mintz2009distant, chen2017reading}.

\begin{itemize}
\setlength{\itemsep}{0pt}
\setlength{\parsep}{0pt}
\setlength{\parskip}{0pt}
\item[1)] \textbf{\EVI} \quad The evidence, such as entities and their corresponding relations, should exist across the query and the targeted passage as they both discuss similar facts or events related to the answer.

\item[2)] \textbf{Answer Containing} \quad The golden passage should contain the answer of the query, which means that a text span within the passage can provide the information-seeking target of the query.

\end{itemize}

In this paper, we propose \textbf{H}yper\textbf{L}ink-induced \textbf{P}re-training (\name), a pre-training method to learn effective Q-P relevance induced by the hyperlink topology within naturally-occurring Web documents. Specifically, these Q-P pairs are automatically extracted from the online documents with relevance adequately designed via hyperlink-based topology to facilitate downstream retrieval for question answering. Figure \ref{Fig:img1} shows an example of comparison between the human-written query and different pseudo queries. By the guidance of hyperlinks, our \name~query hold the relevance of answer containing with the passage~(query title occurs in the passage). Meanwhile, the \name~query can introduce far more effective relevance of \evi~than other pseudo queries by deeply mining the hyperlink topology, e.g., the dual-link structure. In figure \ref{Fig:img1}, both \name~query and the passage both contain information corresponding to the same fact of \textit{``Mitja Okorn directed the film of Letters to Santa''}. This makes our pseudo query low-cost and a good surrogate for the manually written query.

Our contributions are two-fold. First, we present a hyperlink-induced relevance construction methodology that can better facilitate downstream passage retrieval for question answering, and specifically, we propose a pre-training method: Hyperlink-induced Pre-training (\name). Second, we conduct evaluations on six popular QA datasets, investigating the effectiveness of our approach under zero-shot, few-shot, multi-hop, and out-of-domain (OOD) scenarios. The experiments show \name~outperforms BM25 in most of the cases under the zero-shot scenario and other pre-training methods under all scenarios.

\begin{figure*}[tb]
	\begin{center}
		\includegraphics[width=1\textwidth]{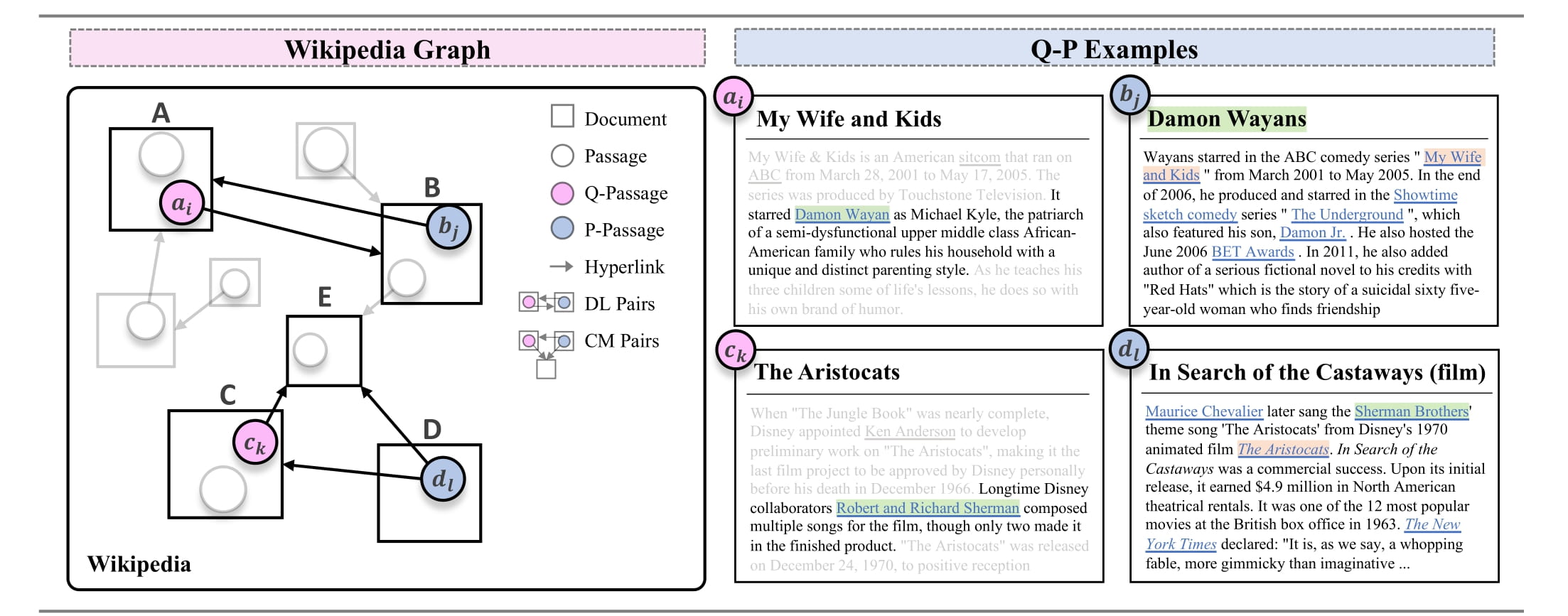}
		\caption{The figure on the left shows a partial Wikipedia graph where two types of pseudo Q-P pairs $(a_i,b_j)$ and $(c_k,d_l)$ are presented. The text boxes on the right show two concrete examples of \name~Q-P pairs where text highlighted in green gives evidence while in orange indicates the answer span.}
		\label{Fig:main}
	\end{center}
\end{figure*}

\section{Related Work}
\textbf{Dense Retriever Pre-training} \quad
Previous works have attempted to conduct additional pre-training for dense retrievers on various weakly supervised data. \citet{web_search_pretrain_1} and \citet{web_search_pretrain_2} pre-trained ranking models on click-logs and BM25-induced signals respectively for web search.
\citet{lee2019latent} proposed the inverse cloze task~(ICT) to pre-train a dense retrieval model, which randomly selects sentences as pseudo queries, and matched them to the passages that they originate from. Besides, \citet{chang2020pre} proposed the pre-training task of wiki link prediction~(WLP) and body first selection~(BFS) tasks. Similar to our work, the WLP task also leveraged the hyperlinks within Wikipedia to construct relevant text pairs. However, as shown in figure~\ref{Fig:img1}, the WLP pseudo query can only ensure the weak doc-wise contextual relationship with the passage. \citet{guu2020retrieval} proposed the masked-salient-span pre-training task which optimizes a retrieval model by the distant supervision of language model objective.
As a follow-up, \citet{sachan2021end} combined ICT with the masked-salient-span task and further improved the pre-training effectiveness.


\noindent \textbf{Data Augmentation via Question Generation} \quad 
\citet{zero-shot-qg}, \citet{reddy2021towards} and \citet{ouguz2021domain} all investigate training a dense retriever on questions synthesized by large question generative~(QG) models. Targeting on the zero-shot setting, \citet{zero-shot-qg} trained a question generator on general-domain question passage pairs from community platforms and publicly available academic datasets. \citet{reddy2021towards} focused more on domain transfer and trained the QG model on QA datasets of Wikipedia articles. \citet{ouguz2021domain} uses the synthetically generated questions from PAQ dataset~\cite{paq} and the post-comment pairs from dataset of Reddit conversations for retrieval pre-training. Recently, \citet{shinoda2021can} reveals that the QG models tend to generate questions with high lexical overlap which amplify the bias of QA dataset. Different to these studies, our method focuses on a more general setting where the retriever is only trained with the naturally occurring web documents, and has no access to any downstream datasets.

\section{Hyperlink-induced Pre-training~(\name)}
\label{sec:methodology}
In this section, we firstly discuss the background of OpenQA retrieval, then our methodology and training framework.

\subsection{Preliminaries}
\noindent \textbf{Passage Retrieval} \quad 
Given a question $q$, passage retrieval aims to provide a set of relevant passages $p$ from a large corpus $\mathcal{D}$. Our work adopts Wikipedia as source corpus and each passage is a disjoint segment within a document from $\mathcal{D}$.

\noindent \textbf{OpenQA Q-P Relevance} \quad 
For OpenQA, a passage $p$ is considered relevant to the query $q$ if $p$ conveys similar facts and contains the answer to $q$. These two conditions of relevance, namely \evi~and answer containing, are properly introduced into the \name~Q-P pairs under the guidance of desired hyperlink structure. We will discuss more in this section.

To better formulate the relevance of pseudo Q-P pairs, we denote the sequence of passages within a document as $A = [a_1, a_2, ..., a_{n_A}]$ where $A \in \mathcal{D}$. The corresponding topical entity and the title of document $A$ and its passage splits are denoted as $e_A$ and $t_A$, respectively. We use $m_A$ to indicate a mention of entity $e_A$, which is a hypertext span linking to document $A$. Note that the mention span $m_A$ is usually identical to the document title $t_A$ or a variant version of it.
Further, we define $\mathcal{F}_{(p)}$ as the entity-level factual information conveyed by the passage $p$, which is a set consists of the topical entity $e_{P}$ and the entities mentioned within passage $p$.

\noindent \textbf{\EVI~in \name} \quad  
With appropriately designed hyperlink topologies, our \name~Q-P pairs guarantee the co-occurrence of entities which are presented as hypertext or topics in $q$ and $p$. This is considered as evidence across the Q-P pairs:

\begin{gather}
 \mathcal{F}_{(q)} \cap \mathcal{F}_{(p)} \neq \emptyset \label{con:evidence0}
\end{gather}

Furthermore, we conjecture that \name~is more likely to achieve fact-level relevance than entity-level overlap. We conduct human evaluation in Section \ref{sec:human} and case studies in Appendix \ref{appendix:case} to support this conjecture. Moreover, we demonstrate that any Q-P pair containing hyperlink-induced factual evidence, which can be represented as triples, is included in our proposed topologies, which are included in Appendix \ref{appendix:reduction}.

\noindent \textbf{Answer Containing in \name} \quad 
We consider the document title $t_Q$ as the information-seeking target of $q$. Accordingly, the relevance of answer containing can be formulated as


\begin{gather}
  t_Q \subseteq p  \label{con:answer0} 
\end{gather}

The rationale behind this is that both the natural question and the Wikipedia document are intended to describe related facts and events regarding a targeted object, whereas the object is an answer for a question but a topical entity for a Wikipedia document. This similarity leads us to take the document title as the information-seeking target of its context.

\subsection{Hyperlink-induced Q-P Pairs}
\label{sec:HIS}
Based on analysis of how queries match their evidential passages in the NQ~\cite{kwiatkowski2019natural} dataset, we propose two kinds of hyperlink topology for relevance construction: Dual-link and Co-mention. We present our exploratory data analysis on NQ dataset in Appendix \ref{appendix:analysis_evidence}. Here we discuss the desired hyperlink topologies and the corresponding relevance of the pseudo Q-P pairs.

\noindent \textbf{Dual-link (DL)}\quad 
Among all NQ training samples, 55\% of questions mention the title of their corresponding golden passage. This observation motivates us to leverage the topology of dual-link~(DL) for relevance construction. We consider a passage pair $(a_i, b_j)$ follows the dual-link topology if they link to each other. An example of a DL pair $(a_i, b_j)$ is shown in Figure \ref{Fig:main}, in which passage $b_j$ mentions the title of document $A$ as $m_A$, satisfying the condition of answer containing:

\begin{gather}
 t_A \approx m_A {\rm \quad and \quad} m_A  \subseteq b_j  \label{con:answer1} 
\end{gather}

\noindent Further, since the passages $a_{i}$ and $b_{j}$ both mention the topical entity of the other, the entities $e_{A}$ and $e_{B}$ appear in both passages as evidence:

\begin{gather}
   \{e_A, e_B\} \subseteq  \mathcal{F}_{(a_i)} \cap \mathcal{F}_{(b_j)} \label{con:evidence1}
\end{gather}

\noindent \textbf{Co-mention (CM)} \quad Among all NQ training samples, about 40\% of questions fail to match the dual-link condition but mention the same third-party entity as their corresponding golden passages. In light of this observation, we utilize another topology of Co-mention~(CM). We consider that a passage pair $(c_k, d_l)$ follows the Co-mention topology if they both link to a third-party document $E$ and $d_l$ links to $c_k$. Figure \ref{Fig:main} illustrates a CM pair $(c_l, d_k)$ where answer containing is ensured as the title of $c_k$ occurs in $d_l$:

\begin{gather}
 t_C \approx m_C {\rm \quad and \quad} m_C  \subseteq d_l \label{con:answer2}
\end{gather}

\noindent Since both $c_l$ and $d_k$ mention a third-party entity $e_E$, and that $e_C$ is a topical entity in $c_l$ while a mentioned entity in $d_k$, we have entity-level evidence across $c_l$ and $d_k$ as:

\begin{gather}
  \{e_C, e_E\}  \subseteq  \mathcal{F}_{(c_k)} \cap \mathcal{F}_{(d_l)} \label{con:evidence2}
\end{gather}

In practice, we use sentence-level queries which contain the corresponding evidential hypertext, and we do not prepend the title to the passage in order to reduce the superficial entity-level overlap. To improve the quality of CM pairs, we filter out those with a co-mentioned entity which has a top 10\% highest-ranked in-degree among the Wikipedia entity. We also present pseudo code in Appendix \ref{appendix:pseudo} to illustrate how we construct our pseudo Q-P pairs.

Furthermore, we highlight that \name~has the following advantages:
1) it introduces more semantic variants and paraphrasing for better text matching. 
2) The hypertext reflects potential interests or needs of users in relevant information, which is consistent to the downstream information-seeking propose.

\subsection{Bi-encoder Training}
We adopt a BERT-based bi-encoder to encode queries and passages separately into d-dimension vectors. The output representation is derived from the last hidden state of the [CLS] token and the final matching score is measured by the inner product:

\begin{align}
 & h_q = {\rm BERT_Q}(q)({\rm[CLS]}) \nonumber \\
 & h_p = {\rm BERT_P}(p)({\rm[CLS]}) \nonumber \\
 & {\rm S}(p,q) = h_q^{\rm T} \cdot h_p  \nonumber
\end{align}

Let $B=\{ \langle q_{i}, p_{i}^{+}, p_{i}^{-} \rangle \}_{i=1}^{n} $ be a mini-batch with $n$ instances. Each instance contains a question $q_{i}$ paired with a positive passage $p_{i}^{+}$ and a negative passage $p_{i}^{-}$. With in-batch negative sampling, each question $q_{i}$ considers all the passages in $B$ except its own gold $p_{i}^{+}$ as negatives, resulting in $2n-1$ negatives per question in total. We use the negative log likelihood of the positive passage as our loss for optimization:

\begin{align}
  &  L(q_{i}, p_{i}^{+}, p_{i,1}^{-},...,p_{i,2n-1}^{-}) \nonumber \\
= &  -log\frac
{e^{S(q_{i},p_{i}^{+})}}
{  e^{S(q_{i},p_{i}^{+})} + 
   {\textstyle \sum_{j=1}^{2n-1}} e^{S(q_{i},p_{i,j}^{-})} 
}   \nonumber
\end{align}

\section{Experimental Setup}
In this session, we discuss the pre-training corpus preparation, downstream datasets, the hyper-parameter and the basic setup for our experiments.

\subsection{Pre-training Corpus}
We adopt Wikipedia as our source corpus $\mathcal{D}$ for pre-training as it is the largest encyclopedia covering diverse topics with good content quality and linking structures. We choose the snapshot 03-01-2021 of an English Wikipedia dump, and process it with WikiExtractor\footnote{Available at https://github.com/attardi/wikiextractor} to obtain clean context. After filtering out documents with blank text or a title less than three letters, following previous work~\cite{karpukhin2020dense}, we split the remaining documents into disjoint chunks of 100 words as passages, resulting in over 22 million passages in the end.

\subsection{Downstream Datasets}
We evaluate our method on several open-domain question answering benchmarks which are shown below.

\noindent \textbf{Natural Questions (NQ)} \cite{kwiatkowski2019natural} is a popular QA dataset with real queries from Google Search and annotated answers from Wikipedia.

\noindent \textbf{TriviaQA} \cite{joshi2017triviaqa} contains question-answer pairs scraped from trivia websites. 

\noindent \textbf{WebQuestions (WQ)} \cite{berant2013semantic} consists of questions generated by Google Suggest API with entity-level answers from Freebase.

\noindent \textbf{HotpotQA} (Fullwiki) \cite{yang2018hotpotqa} is a human-annotated multi-hop question answering dataset.

\noindent \textbf{BioASQ} \cite{tsatsaronis2015overview} is a competition on biomedical semantic indexing and question answering. We evaluate its factoid questions from task 8B. 

\noindent \textbf{MS MARCO} (Passage Ranking) \cite{nguyen2016ms} consists of real-world user queries and a large collection of Web passages extracted by Bing search engine.

\noindent \textbf{Retrieval Corpus} \quad For downstream retrieval, we use the 21M Wikipedia passages provided by DPR \cite{karpukhin2020dense} for NQ, TriviaQA and WQ. For BioASQ, we take the abstracts of PubMed articles from task 8A with the same split to \citet{reddy2021towards}'s work. For HotpotQA and MS MARCO, we use the official corpus.

\subsection{Implementation Details}
During the pre-training, we train the bi-encoder for 5 epochs with parameters shared, using a batch size of 400 and an Adam optimizer \citep{kingma2014adam} with a learning rate $2 \times 10^{-5}$, linear scheduling with 10\% warm-up steps. Our \name~and all the reproduced baselines are trained on 20 million Q-P pairs with in-batch negative sampling, and the best checkpoints are selected based on the average rank of gold passages evaluated on the NQ dev set. The pre-training takes around 3 days using eight NVIDIA V100 32GB GPUs. 

For the downstream, we use the same hyper-parameters for all experiments. Specifically, we fine-tune the pre-trained models for 40 epochs with a batch size of 256 and the same optimizer and learning rate settings to the pre-training. We conduct evaluation on respective dev sets to select best checkpoints, and we use the last checkpoint if there is no dev set or test set (e.g. HotpotQA). More details can be found in the Appendix \ref{appendix:parameter}.

\subsection{Baselines}
Most existing baselines have been implemented under different experimental settings,  which have a substantial effect on the retrieval performance. To ensure fairness, we reproduce several pre-training methods (ICT, WLP, BFS, and their combination) under the same experimental setting, such as batch size, base model, amount of pre-training data, and so on. 
The only difference between our method and the re-implemented baselines is the self-supervision signal derived from the respective pre-training samples. Our reproduced BM25 baseline is better than that reported in \citet{karpukhin2020dense}, and the re-implemented pre-training methods also perform better than those reported by the recent work\footnote{Our reproduced ICT and BFS surpass the reproduction from recent work \cite{ouguz2021domain} by 15 and 12 points, respectively, in terms of top-20 retrieval accuracy on NQ test set under zero-shot setting.}.
In addition, we include the work REALM \cite{guu2020retrieval} as a baseline which has recently been reproduced by \citet{sachan2021end} using 240 GPUs and is named masked salient spans (MSS). 
We note that most related works gain improvements from varying downstream setting or synthetic pre-training with access to the downstream data of respective domain, which is out of the scope of our interests.

\begin{table*}
\centering
\scalebox{0.8}{
\begin{tabular}{lccccccccc} 
\hline
             & \multicolumn{3}{c}{NQ}                        & \multicolumn{3}{c}{TriviaQA}                  & \multicolumn{3}{c}{WQ}                         \\
             & top5          & top20         & top100        & top5          & top20         & top100        & top5          & top20         & top100         \\ 
\hline
             & \multicolumn{9}{c}{w/o fine-tuning (zero-shot)}                                                                                                            \\ 
\cline{2-10}
$\textrm{BM25}^{\dag}$   & 43.6          & 62.9          & 78.1          & \textbf{66.4} & 76.4          & 83.2          & 42.6          & 62.8          & 76.8           \\
$\textrm{ICT}^{\dag}$ \cite{lee2019latent}  & 23.4          & 40.7          & 58.1          & 33.3          & 51.3          & 69.9          & 19.9          & 36.2          & 56.0           \\
$\textrm{WLP}^{\dag}$ \cite{chang2020pre}          & 28.5          & 47.3          & 65.3          & 51.3          & 67.0          & 79.1          & 26.9          & 49.0          & 68.1           \\
$\textrm{BFS}^{\dag}$ \cite{chang2020pre}          & 31.0          & 49.9          & 67.5          & 43.8          & 61.1          & 74.7          & 28.5          & 48.0          & 67.7           \\
$\textrm{ICT+WLP+BFS}^{\dag}$ \cite{chang2020pre}  & 32.3          & 50.2          & 68.0          & 49.7          & 65.5         & 78.3          & 28.4          & 47.8          & 67.5          \\
MSS \cite{sachan2021end}     & 41.7          & 59.8          & 74.9          & 53.3          & 68.2          & 79.4          & -             & -             & -              \\
\name    & \textbf{51.2} & \textbf{70.2} & \textbf{82.0} & 65.9          & \textbf{76.9} & \textbf{84.0} & \textbf{49.3} & \textbf{66.9} & \textbf{80.8}  \\ 
\hline
             & \multicolumn{9}{c}{w/ fine-tuning}                                                                                                             \\ 
\cline{2-10}
$ \textrm{No Pre-train}^{\dag}$  & 68.5          & 79.6          & 86.5          & 71.3          & 79.7          & 85.0          & 61.6          & 74.5          & 81.7           \\
$\textrm{ICT}^{\dag}$ \cite{lee2019latent}  & 69.8          & 81.1          & 87.0          & 70.4          & 79.8          & 85.5          & 63.7          & 75.5          & 83.4           \\
$\textrm{WLP}^{\dag}$ \cite{chang2020pre}  & 69.8          & \textbf{81.4}          & 87.4          & 73.1          & 81.5          & 86.1          & 64.5          & 75.2          & 83.9           \\
$\textrm{BFS}^{\dag}$ \cite{chang2020pre}  & 68.7          & 80.1          & 86.5          & 72.8          & 80.8          & 86.0          & 63.0          & 75.1          & 83.5           \\
$\textrm{ICT+WLP+BFS}^{\dag}$ \cite{chang2020pre}  & 68.9          & 80.9          & 87.7          & 74.6          & 82.2         & 86.5          & 64.1          & \textbf{76.7}          & 84.4          \\
\name    & \textbf{70.9} & \textbf{81.4}          & \textbf{88.0} & \textbf{75.3} & \textbf{82.4} & \textbf{86.9} & \textbf{65.5} & 76.5 & \textbf{84.5}  \\
\hline
\end{tabular}
}
\caption{Top-k $(k \in \{5,20,100\})$ retrieval accuracy, measured as the percentage of top $k$ retrieved passages with the answer contained. The upper block of the table describes the performance under zero-shot setting, while the lower under the full-set fine-tuning setting. $\dag$: Our re-implementation.}
\label{table:main-result}
\vspace{-0.2cm} 
\end{table*}

\section{Experiments}
\subsection{Main Results}
Table \ref{table:main-result} shows the retrieval accuracy of different models on three popular QA datasets under zero-shot and full-set fine-tuning settings. 

Under zero-shot setting, \name~consistently outperforms BM25 except for the top-5 retrieval accuracy of TriviaQA, while all other pre-training baselines are far behind. 
We attribute the minor improvement over BM25 on TriviaQA to a high overlap between questions and passages, which gives term-based retriever a clear advantage. We investigate the coverage of the question tokens that appear in the gold passage and find that the overlap is indeed higher in TriviaQA (62.8\%) than NQ (60.7\%) and WQ (57.5\%). 

After fine-tuning, all models with intermediate pre-training give better results than the vanilla DPR while our \name~achieves the best in nearly all cases. 
Among ICT, WLP and BFS, we observe that WLP is the most competitive with or without fine-tuning, and additional improvements can be achieved by combining three of them. 
This observation indicates that pre-training with diverse relevance leads to better generalization to downstream tasks, while document-wise relevance is more adaptable for the OpenQA retrieval.
The advantage of document-wise relevance may come from the fact that texts in different documents are likely written by different parties, providing less superficial cues for text matching, which is beneficial for the downstream retrieval. 
Our \name~learns both coarse-grained document-wise relationships as well as the fine-grained entity-level evidence, which results in a significant improvement.

\subsection{Few-shot Learning}

To investigate the retrieval effectiveness in a more realistic scenario, we conduct experiments for few-shot learning.
Specifically, we fine-tune the pre-trained models on large datasets (NQ, TriviaQA) with $m$ ($m\in\{16, 256, 1024\}$) samples and present the few-shot retrieval results in Table \ref{table:few-shot}. 
With only a few hundred labeled data for fine-tuning, 
all the models with intermediate pre-training perform better than those without, and \name~outperforms the others by a larger margin when $m$ is smaller. 
Moreover, among three re-implemented baselines, WLP gains the largest improvement with increasing number of samples, outperforming ICT and BFS when a thousand labelled samples are provided for fine-tuning. 

\begin{table}[H]
\centering
\scalebox{0.7}{
\centering
\begin{tabular}{lcccccc} 
\hline
             & \multicolumn{3}{c}{NQ}                        & \multicolumn{3}{c}{TriviaQA}                     \\
             & top5          & top20         & top100        & top5          & top20         & top100         \\ 
\hline
             & \multicolumn{6}{c}{m = 16}                                                      \\ 
\cline{2-7}
No Pre-train & 12.7          & 24.2          & 40.2          & 18.6          & 32.6          & 51.0           \\
ICT          & 37.1          & 54.4          & 70.5          & 47.2          & 62.5          & 75.8           \\
WLP          & 29.8          & 48.2          & 65.5          & 51.4          & 66.9          & 79.2           \\
BFS          & 39.8          & 57.9          & 73.2          & 46.9          & 62.2          & 75.2           \\
\name          & \textbf{51.9} & \textbf{70.3} & \textbf{81.6} & \textbf{65.9} & \textbf{76.9} & \textbf{84.0}  \\ 
\hline
             & \multicolumn{6}{c}{m = 128}                                                     \\ 
\cline{2-7}
No Pre-train & 38.0          & 53.4          & 68.8          & 38.0          & 53.4          & 68.8           \\
ICT          & 47.0          & 64.2          & 77.4          & 58.5          & 71.4          & 81.0           \\
WLP          & 44.9          & 62.4          & 76.6          & 63.1          & 74.5          & 82.6           \\
BFS          & 44.4          & 62.8          & 76.7          & 59.2          & 71.7          & 80.8           \\
\name          & \textbf{55.2} & \textbf{71.3} & \textbf{81.8} & \textbf{67.7} & \textbf{77.7} & \textbf{84.4}  \\ 
\hline
             & \multicolumn{6}{c}{m = 1024}                                                    \\ 
\cline{2-7}
No Pre-train & 49.7          & 66.4          & 78.8          & 54.0          & 67.2          & 77.6           \\
ICT          & 55.9          & 72.2          & 83.7          & 63.8          & 75.7          & 83.3           \\
WLP          & 57.2          & 73.6          & 83.9          & 67.2          & 77.5          & 84.5           \\
BFS          & 53.7          & 71.7          & 83.1          & 63.6          & 75.3          & 83.1           \\
\name         & \textbf{60.6} & \textbf{76.4} & \textbf{85.3} & \textbf{70.2} & \textbf{79.8} & \textbf{85.4}  \\
\hline
\end{tabular}}
\caption{Few-shot retrieval accuracy on NQ and TriviaQA test sets after fine-tuning with $m$ annotated samples.}
\label{table:few-shot}
\end{table}

\subsection{Out-of-domain (OOD) Scenario}
\label{sec:ood}

\begin{table*}[]
\centering
\scalebox{0.8}{
\begin{tabular}{lcccccccccc} 
\hline
\multirow{2}{*}{Model}      & \multirow{2}{*}{Negative} & \multicolumn{3}{c}{NQ}                           & \multicolumn{3}{c}{TriviaQA}                     & \multicolumn{3}{c}{WebQ}                          \\
                            &                           & top5           & top20          & top100         & top5           & top20          & top100         & top5           & top20          & top100          \\ 
\hline
\multirow{2}{*}{Dual-link}  & 0                         & 46.2           & 64.7           & 78.0           & 60.5           & 73.0           & 81.2           & 44.6           & 65.2           & 78.8            \\
                            & 1                         & 49.0           & 67.8           & 79.7           & 62.0           & 73.8           & 82.1           & 48.4           & \textbf{67.1}           & 79.5            \\ 
\hline
\multirow{2}{*}{Co-mention} & 0                         & 35.8           & 57.1           & 75.1           & 58.9           & 73.1           & 82.6           & 36.2           & 58.9           & 76.2            \\
                            & 1                         & 42.5           & 62.2           & 77.9           & 63.2           & 75.8           & 83.7           & 45.4           & 64.5           & 78.9            \\ 
\hline
\multirow{2}{*}{\name}           & 0                         & 45.7           & 66.0           & 79.9           & 62.6           & 75.2           & 83.0           & 43.9           & 64.1           & 79.4            \\
                            & 1                         & \textbf{51.2} & \textbf{70.2} & \textbf{82.0} & \textbf{65.9} & \textbf{76.9} & \textbf{84.0} & \textbf{49.3} & 66.9 & \textbf{80.8}  \\
\hline
\end{tabular}}
\caption{Ablation studies on different types of topologies and negatives. The retrieval accuracy of models trained with different types of Q-P pairs and additional negatives on NQ, TriviaQA, and WebQ datasets.}
\label{table:abl}
\vspace{-0.3cm} 
\end{table*}

While \name~is pre-trained on Wikipedia pages, we conduct additional experiments on BioASQ and MS MARCO datasets with non-Wikipedia corpus to further verify its out-of-domain (OOD) generalization.
Following \citet{gururangan2020don}, we measure the similarity between corpus by computing the vocabulary overlap of the top 10K frequent words (excluding stopwords). We observe a vocabulary overlap of 36.2\% between BioASQ and Wikipedia while 61.4\% between MS MARCO and Wikipedia, indicating that these two domains differ considerably from our pre-training corpus.

The results of zero-shot retrieval on BioASQ and MS MARCO datasets are presented in Table \ref{table:ood}. 
For BioASQ, \name~is competitive with both BM25 and AugDPR\cite{reddy2021towards} while significantly outperforming ICT, WLP, and BFS. Note that AugDPR is a baseline that has access to NQ labeled data whereas our \name~is trained in an unsupervised way. 
For MS MARCO, \name~consistently outperforms other pre-training methods but falls behind BM25 under zero-shot setting. We conjecture the performance degradation on MS MARCO is attributed to two factors: 1) the Q-P lexical overlap of MS MARCO (65.7\%) is higher than that in BioASQ (48.7\%) as well as other datasets; 2) the information-seeking target of the MS MARCO query is the entire passage rather than a short answer span, which is biased towards our proposed answer containing. 
we also observe that pre-training exclusively with DL pairs achieves better results in MS MARCO, indicating the generality of relevance induced by DL topology.

\begin{table}[H]
\centering
\scalebox{0.8}{
\begin{tabular}{lcccc} 
\hline
            & \multicolumn{2}{c}{BioASQ}    & \multicolumn{2}{c}{MS MARCO}   \\
            & top20         & top100        & R@20          & R@100          \\ 
\hline
BM25        & $\textrm{42.1}^{\ddag}$   & $\textrm{50.5}^{\ddag}$         & \textbf{49.0} & \textbf{69.0}  \\
DPR         & $\textrm{34.7}^{\ddag}$   & $\textrm{46.9}^{\ddag}$        & -             & -              \\
AugDPR      & $\textrm{41.4}^{\ddag}$   & $\textrm{52.4}^{\ddag}$          & -             & -              \\ 
\hline
ICT         & 8.9           & 18.6          & 10.8          & 19.5           \\
WLP         & 29.7          & 44.3          & 18.4          & 36.0           \\
BFS         & 28.4          & 41.9          & 28.0          & 44.7           \\ 
\hline
\name~(DL)    & \textbf{46.0} & 56.9         & 42.0        & 62.6              \\
\name~(CM)    & 37.8          & 54.7          & 26.6       & 47.3              \\
\name~(DL+CM) & 40.8          & \textbf{58.3} & 37.3          & 60.0           \\
\hline
\end{tabular}}
\caption{Top-20/100 zero-shot retrieval accuracy on BioASQ and Top-20/100 zero-shot recall on MS MARCO. \ddag: \cite{reddy2021towards}}
\label{table:ood}
\vspace{-0.2cm}
\end{table}

\subsection{Multi-hop Retrieval}
While \name~aims to acquires the ability in matching document-wise concepts and facts, it raises our interest in its capability for multi-hop scenarios. We evaluate our methods on HotpotQA in a single-hop manner. Specifically, for each query, we randomly selects one golden passage from the two as a positive passage and one additional passage with high TF-IDF scores as a negative passage. 
Our models are further fine-tuned on the HotpotQA training set and evaluated on the bridge and the comparison type questions from the development set, respectively.
The results of our study are shown in Table \ref{table:hotpot} which reveals that \name~consistently outperforms others methods, with up to a 11-point improvement on top-5 retrieval accuracy of bridge questions. Furthermore, WLP yields a 4-point advantages in average over ICT and BFS on bridge questions, showing that document-wise relevance contributes to better associative abilities. We include a case study in Appendix \ref{appendix:hotpot}. 

\begin{table}[H]
\centering
\scalebox{0.7}{
\centering
\begin{tabular}{lcccccc} 
\hline
\multirow{2}{*}{} & \multicolumn{3}{c}{Bridge}                       & \multicolumn{3}{c}{Comparison}                    \\
                  & top5           & top20          & top100         & top5           & top20          & top100          \\ 
\hline
No Pre-train      & 25.0           & 40.5           & 58.0           & 83.0           & 94.2           & 97.4            \\
ICT               & 28.1           & 43.8           & 61.8           & 84.8           & 94.4           & 98.3            \\
WLP               & 32.1           & 49.1           & 66.0           & 89.7           & 97.3           & 99.2            \\
BFS               & 29.0           & 44.7           & 62.1           & 87.4           & 95.8           & 98.7            \\
\name         & \textbf{36.9} & \textbf{53.0} & \textbf{68.5} & \textbf{94.4} & \textbf{98.5} & \textbf{99.5}  \\
\hline
\end{tabular}
}
\caption{Retrieval accuracy on questions from HotpotQA dev set, measured as the percentage of top-k retrieved passages which include both golds.}
\label{table:hotpot}
\vspace{-0.2cm}
\end{table}

\section{Analysis}

\subsection{Ablation Study}
\label{sec:abl}
To better understand how different key factors affect the results, we conduct ablation experiments with results shown in Table \ref{table:abl}.

\noindent \textbf{Hyperlink-based Topologies} \quad 
Our proposed dual-link (DL) and co-mention (CM) Q-P pairs, provide evidence induced by different hyperlink-based topologies. 
To examine their respective effectiveness, we pre-train retrievers on Q-P pairs derived from each topology and their combinations. We present zero-shot retrieval results in Table \ref{table:abl}, which show that retrievers pre-trained on DL pairs has a distinct advantage over that on CM pairs, while combining both gives extra improvement. 

\noindent \textbf{Negative Passage} \quad 
In practice, negative sampling is essential for learning a high-quality encoder. Besides in-batch negative, our reported \name~employs one additional negative for each query. We further explore the impact of the additional negatives during pre-training. In our ablation study, pre-training with additional negatives improves the results significantly, which may be attributed to using more in-batch pairs for text matching.
More details on implementation and negative sampling strategies can be found in Appendix \ref{appendix:negative}.

\subsection{Analysis on Q-P Overlap}
We carry out extensive analysis on the Q-P lexical overlap in the task of retrieval. Specifically, we tokenize $q$, $p$ using the BERT tokenizer and measure the Q-P overlap as the proportion of the question tokens that appear in the corresponding passage. Based on the degree of Q-P overlap, we divided the NQ dev set into five categories for further analysis.

\begin{figure}[htbp]
	\centering
	\begin{minipage}[t]{0.23\textwidth}
    \centering
    \includegraphics[width=1\textwidth]{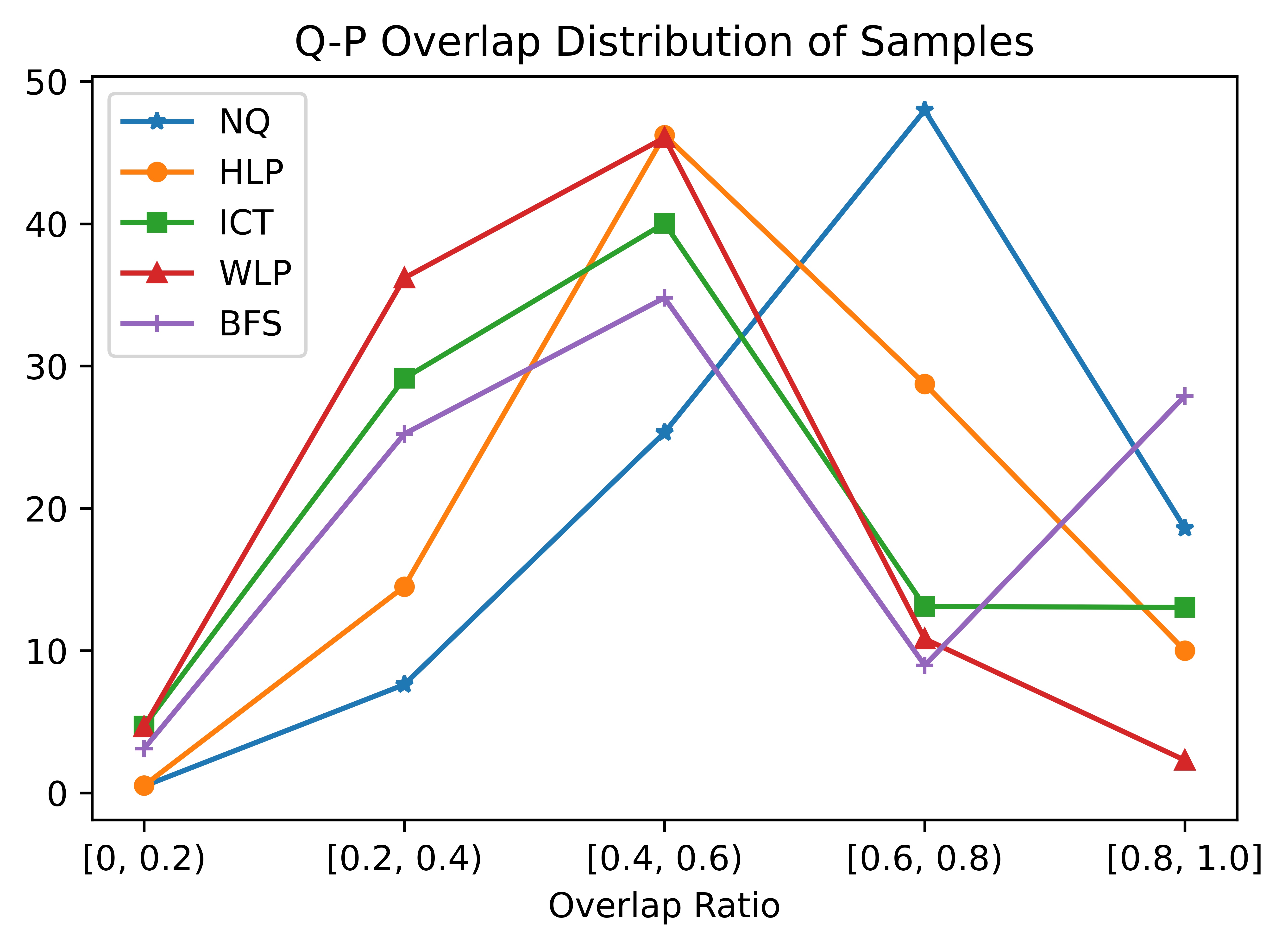}
    \end{minipage}
    \begin{minipage}[t]{0.23\textwidth}
    \centering
    \includegraphics[width=1\textwidth]{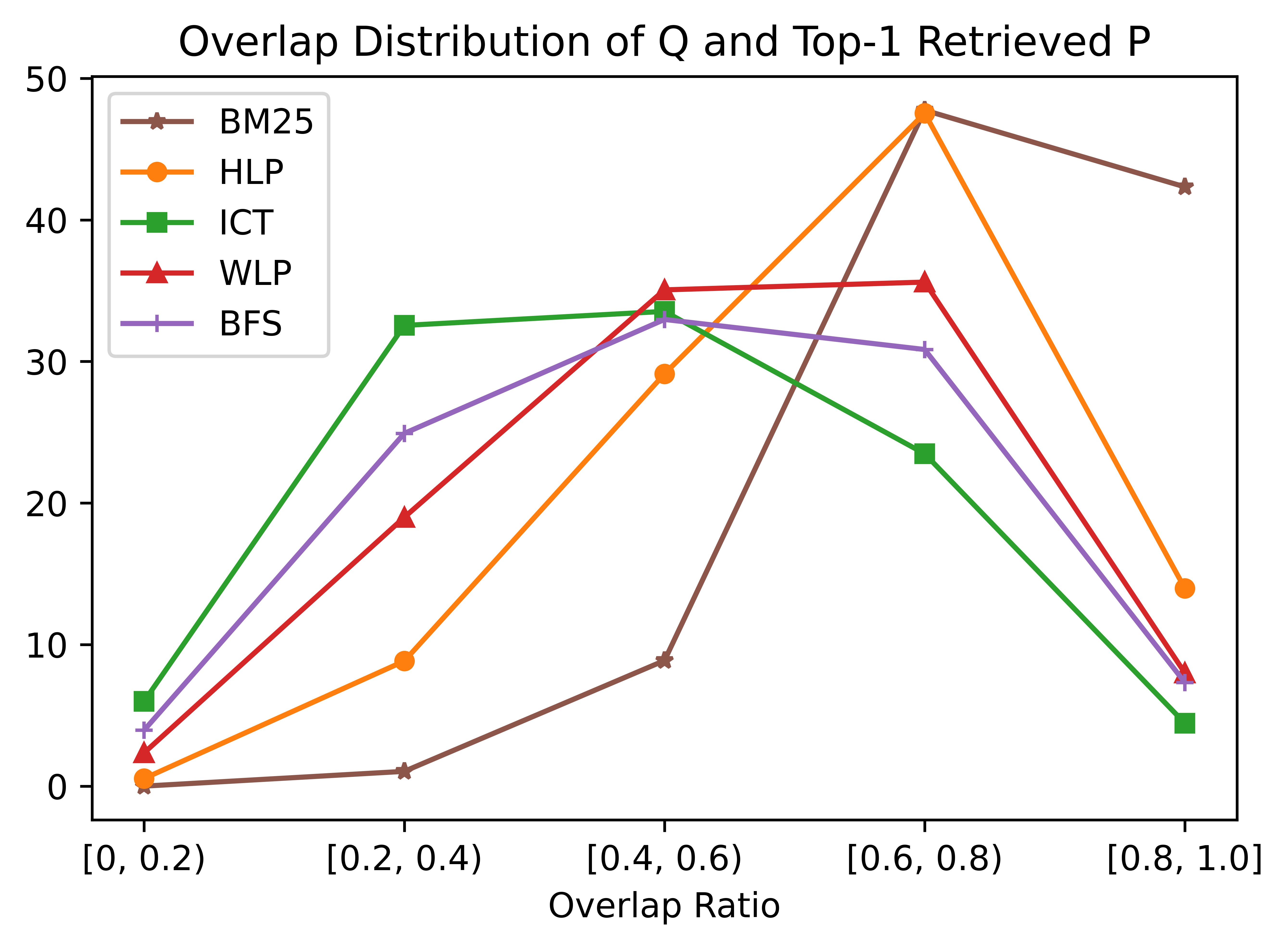}
    \end{minipage}
	\caption{Distribution of overlap on pseudo and downstream Q-P pairs (left), and that between the query and the top-1 passage retrieved by different pre-trained models (right).}
	\label{Fig:overlap-data}
\vspace{-0.2cm}
\end{figure}

\noindent \textbf{Distribution of Q-P Overlap} \quad 
Figure \ref{Fig:overlap-data} shows both the pre-training and the retrieved pairs of \name~have a more similar overlap distribution with the downstream NQ dataset than the other methods, which implies the consistency between the relevance provided by \name~and that in real information-seeking scenario.


\begin{figure}[htbp]
	\centering
	\begin{minipage}[t]{0.23\textwidth}
    \centering
    \includegraphics[width=1\textwidth]{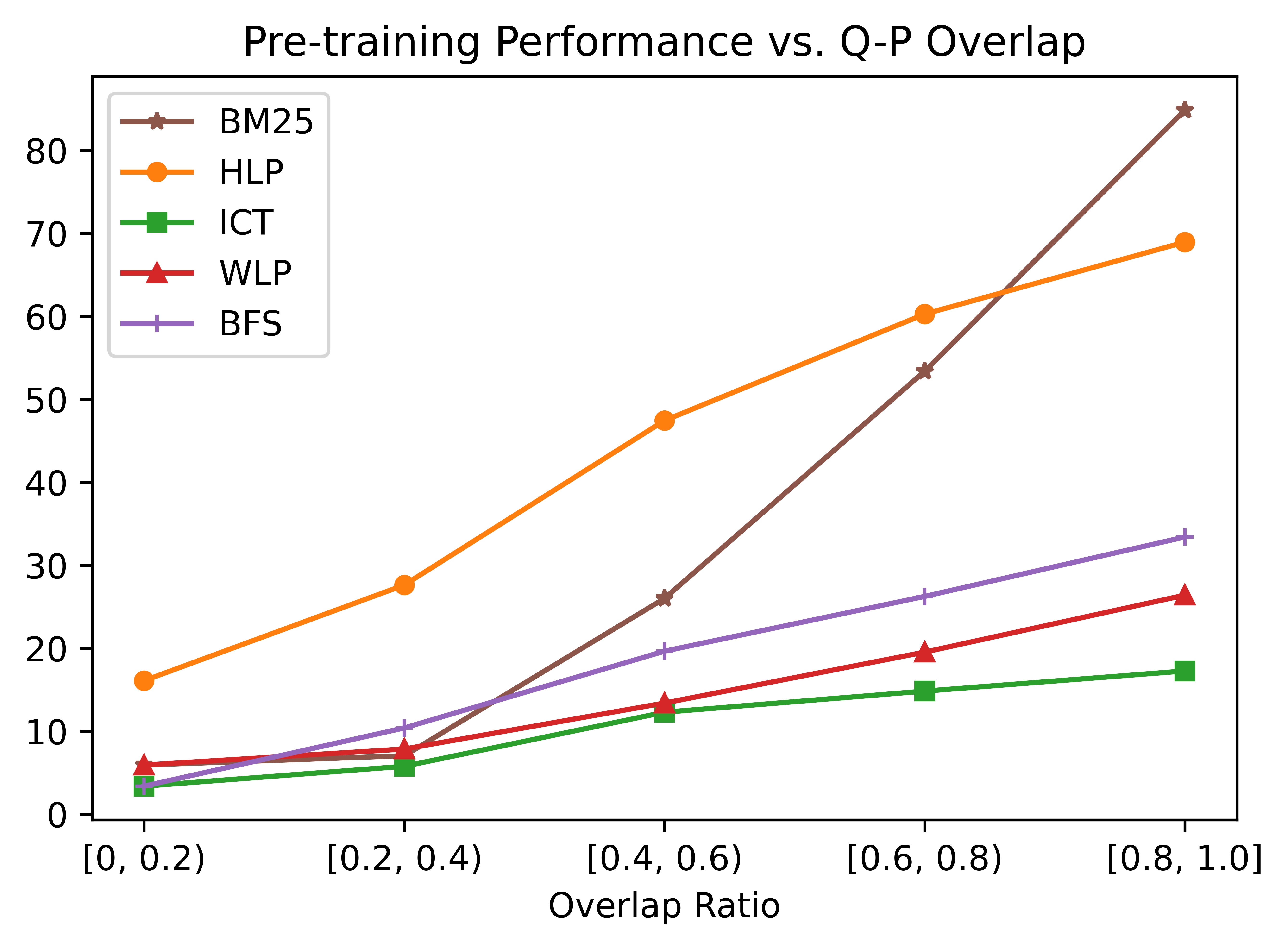}
    \end{minipage}
    \begin{minipage}[t]{0.23\textwidth}
    \centering
    \includegraphics[width=1\textwidth]{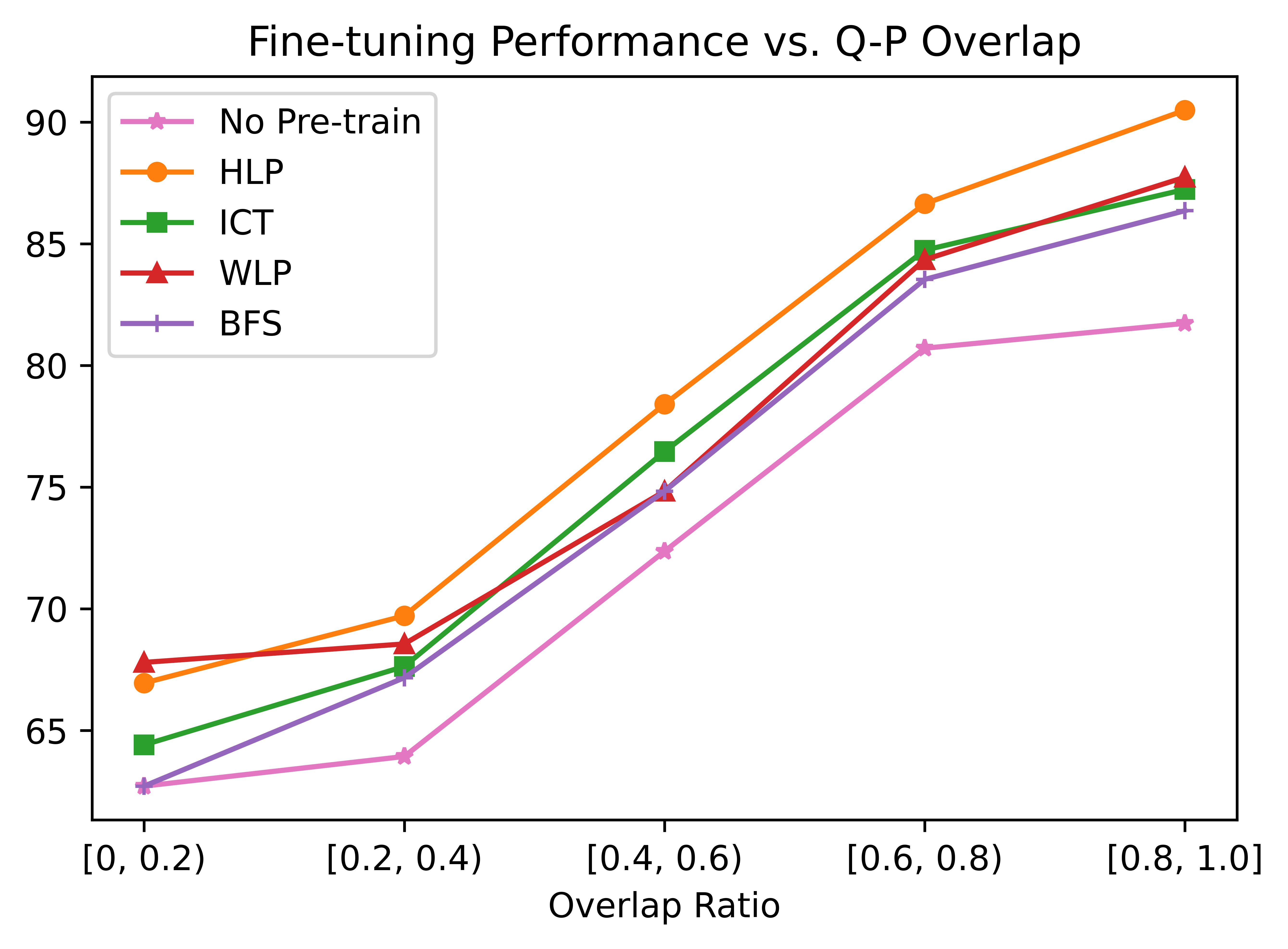}
    \end{minipage}
	\caption{Top-20 retrieval accuracy of pre-training (left) and fine-tuning (right) on the divided NQ dev set.}
	\label{Fig:overlap-performance}
\vspace{-0.2cm}
\end{figure}

\noindent \textbf{Retrieval Performance vs. Q-P Overlap} \quad 
Figure~\ref{Fig:overlap-performance} shows the top-20 retrieval accuracy on the samples with varying degrees of Q-P overlap.
Both figures show that the retrievers are more likely to return answer-containing passages when there is higher Q-P overlap, suggesting that all these models can exploit lexical overlap for passage retrieval. Under the zero-shot setting, \name~outperforms all the methods except BM25 when $r$ is larger than 0.8, which reflects the strong reasoning ability of \name~and the overlap-dependent nature of the term-based retrievers. After fine-tuning, models with additional pre-training perform better than the vanilla DPR while \name~outperforms all other methods in most of the cases. It is important to note that \name~is pre-trained on more high-overlap text pairs while it performs better than all the other methods when fewer overlaps are provided. We speculate that this is because the overlapping in \name~Q-P pairs mostly comes from the factual information, such as entity, which introduces fewer superficial cues, allowing for better adaptation to the downstream cases.

\subsection{Human Evaluation on Q-P pairs}
\label{sec:human}
We conduct human evaluation to investigate the proportion of Q-P pairs that convey the similar fact-level information. Specially, we randomly selected one hundred examples from our constructed Q-P pairs and asked annotators to identify whether the query and the corresponding passage convey similar facts. Each case is evaluated by three annotators and the result is determined by their votes. Our results are shown in Table \ref{table:human}, and we further present case studies in Appendix \ref{appendix:case}.

\begin{table}[H]
\centering
\scalebox{0.8}{
\begin{tabular}{cccc}
\hline
            & DL    & CM    & WLP       \\ \hline
Votes       & 61\%  & 40\%  & 15\%      \\ \hline
\end{tabular}}
\caption{Human evaluation on pseudo Q-P pairs constructed by different methods.}
\label{table:human}
\vspace{-0.2cm}
\end{table}

\section{Conclusion}
This paper proposes Hyperlink-induced Pre-training (\name), a pre-training method for OpenQA passage retrieval by leveraging the online textual relevance induced by hyperlink-based topology. Our experiments show that \name~gains significant improvements across multiple QA datasets under different scenarios, consistently outperforming other pre-training methods. Our method provides insights into OpenQA passage retrieval by analyzing the underlying bi-text relevance. Future work involves addressing tasks like MS MARCO where the granularity of the information-seeking target is at the passage level.

\section{Acknowledgments}
This work is partially supported by the Hong Kong RGC GRF Project 16202218, CRF Project C6030-18G, C1031-18G, C5026-18G, AOE Project AoE/E-603/18, China NSFC No. 61729201. We thank all the reviewers for their insightful comments.

\bibliography{anthology,custom}

\begin{thebibliography}{26}
\expandafter\ifx\csname natexlab\endcsname\relax\def\natexlab#1{#1}\fi

\bibitem[{Berant et~al.(2013)Berant, Chou, Frostig, and
  Liang}]{berant2013semantic}
Jonathan Berant, Andrew Chou, Roy Frostig, and Percy Liang. 2013.
\newblock \href {https://aclanthology.org/D13-1160} {Semantic parsing on
  {F}reebase from question-answer pairs}.
\newblock In \emph{Proceedings of the 2013 Conference on Empirical Methods in
  Natural Language Processing}, pages 1533--1544, Seattle, Washington, USA.
  Association for Computational Linguistics.

\bibitem[{Borisov et~al.(2016)Borisov, Markov, de~Rijke, and
  Serdyukov}]{web_search_pretrain_1}
Alexey Borisov, Ilya Markov, Maarten de~Rijke, and Pavel Serdyukov. 2016.
\newblock A neural click model for web search.
\newblock In \emph{Proceedings of the 25th International Conference on World
  Wide Web, {WWW} 2016, Montreal, Canada, April 11 - 15, 2016}, pages 531--541.

\bibitem[{Chang et~al.(2019)Chang, Felix, Chang, Yang, and
  Kumar}]{chang2020pre}
Wei-Cheng Chang, X~Yu Felix, Yin-Wen Chang, Yiming Yang, and Sanjiv Kumar.
  2019.
\newblock Pre-training tasks for embedding-based large-scale retrieval.
\newblock In \emph{International Conference on Learning Representations}.

\bibitem[{Chen et~al.(2017)Chen, Fisch, Weston, and Bordes}]{chen2017reading}
Danqi Chen, Adam Fisch, Jason Weston, and Antoine Bordes. 2017.
\newblock \href {https://doi.org/10.18653/v1/P17-1171} {Reading {W}ikipedia to
  answer open-domain questions}.
\newblock In \emph{Proceedings of the 55th Annual Meeting of the Association
  for Computational Linguistics (Volume 1: Long Papers)}, pages 1870--1879,
  Vancouver, Canada. Association for Computational Linguistics.

\bibitem[{Dehghani et~al.(2017)Dehghani, Zamani, Severyn, Kamps, and
  Croft}]{web_search_pretrain_2}
Mostafa Dehghani, Hamed Zamani, Aliaksei Severyn, Jaap Kamps, and W.~Bruce
  Croft. 2017.
\newblock Neural ranking models with weak supervision.
\newblock In \emph{Proceedings of the 40th International {ACM} {SIGIR}
  Conference on Research and Development in Information Retrieval, Shinjuku,
  Tokyo, Japan, August 7-11, 2017}, pages 65--74.

\bibitem[{Ferragina and Scaiella(2010)}]{ferragina2010tagme}
Paolo Ferragina and Ugo Scaiella. 2010.
\newblock Tagme: on-the-fly annotation of short text fragments (by wikipedia
  entities).
\newblock In \emph{Proceedings of the 19th ACM international conference on
  Information and knowledge management}, pages 1625--1628.

\bibitem[{Gururangan et~al.(2020)Gururangan, Marasovi{\'c}, Swayamdipta, Lo,
  Beltagy, Downey, and Smith}]{gururangan2020don}
Suchin Gururangan, Ana Marasovi{\'c}, Swabha Swayamdipta, Kyle Lo, Iz~Beltagy,
  Doug Downey, and Noah~A. Smith. 2020.
\newblock \href {https://doi.org/10.18653/v1/2020.acl-main.740} {Don{'}t stop
  pretraining: Adapt language models to domains and tasks}.
\newblock In \emph{Proceedings of the 58th Annual Meeting of the Association
  for Computational Linguistics}, pages 8342--8360, Online. Association for
  Computational Linguistics.

\bibitem[{Guu et~al.(2020)Guu, Lee, Tung, Pasupat, and
  Chang}]{guu2020retrieval}
Kelvin Guu, Kenton Lee, Zora Tung, Panupong Pasupat, and Mingwei Chang. 2020.
\newblock Retrieval augmented language model pre-training.
\newblock In \emph{International Conference on Machine Learning}, pages
  3929--3938. PMLR.

\bibitem[{Joshi et~al.(2017)Joshi, Choi, Weld, and
  Zettlemoyer}]{joshi2017triviaqa}
Mandar Joshi, Eunsol Choi, Daniel Weld, and Luke Zettlemoyer. 2017.
\newblock \href {https://doi.org/10.18653/v1/P17-1147} {{T}rivia{QA}: A large
  scale distantly supervised challenge dataset for reading comprehension}.
\newblock In \emph{Proceedings of the 55th Annual Meeting of the Association
  for Computational Linguistics (Volume 1: Long Papers)}, pages 1601--1611,
  Vancouver, Canada. Association for Computational Linguistics.

\bibitem[{Karpukhin et~al.(2020)Karpukhin, Oguz, Min, Lewis, Wu, Edunov, Chen,
  and Yih}]{karpukhin2020dense}
Vladimir Karpukhin, Barlas Oguz, Sewon Min, Patrick Lewis, Ledell Wu, Sergey
  Edunov, Danqi Chen, and Wen-tau Yih. 2020.
\newblock \href {https://doi.org/10.18653/v1/2020.emnlp-main.550} {Dense
  passage retrieval for open-domain question answering}.
\newblock In \emph{Proceedings of the 2020 Conference on Empirical Methods in
  Natural Language Processing (EMNLP)}, pages 6769--6781, Online. Association
  for Computational Linguistics.

\bibitem[{Kingma and Ba(2014)}]{kingma2014adam}
Diederik~P Kingma and Jimmy Ba. 2014.
\newblock Adam: A method for stochastic optimization.
\newblock \emph{arXiv preprint arXiv:1412.6980}.

\bibitem[{Kwiatkowski et~al.(2019)Kwiatkowski, Palomaki, Redfield, Collins,
  Parikh, Alberti, Epstein, Polosukhin, Devlin, Lee
  et~al.}]{kwiatkowski2019natural}
Tom Kwiatkowski, Jennimaria Palomaki, Olivia Redfield, Michael Collins, Ankur
  Parikh, Chris Alberti, Danielle Epstein, Illia Polosukhin, Jacob Devlin,
  Kenton Lee, et~al. 2019.
\newblock Natural questions: a benchmark for question answering research.
\newblock \emph{Transactions of the Association for Computational Linguistics},
  7:453--466.

\bibitem[{Lee et~al.(2019)Lee, Chang, and Toutanova}]{lee2019latent}
Kenton Lee, Ming-Wei Chang, and Kristina Toutanova. 2019.
\newblock \href {https://doi.org/10.18653/v1/P19-1612} {Latent retrieval for
  weakly supervised open domain question answering}.
\newblock In \emph{Proceedings of the 57th Annual Meeting of the Association
  for Computational Linguistics}, pages 6086--6096, Florence, Italy.
  Association for Computational Linguistics.

\bibitem[{Lewis et~al.(2021)Lewis, Wu, Liu, Minervini, K{\"{u}}ttler, Piktus,
  Stenetorp, and Riedel}]{paq}
Patrick S.~H. Lewis, Yuxiang Wu, Linqing Liu, Pasquale Minervini, Heinrich
  K{\"{u}}ttler, Aleksandra Piktus, Pontus Stenetorp, and Sebastian Riedel.
  2021.
\newblock {PAQ:} 65 million probably-asked questions and what you can do with
  them.
\newblock \emph{CoRR}, abs/2102.07033.

\bibitem[{Ma et~al.(2021)Ma, Korotkov, Yang, Hall, and McDonald}]{zero-shot-qg}
Ji~Ma, Ivan Korotkov, Yinfei Yang, Keith~B. Hall, and Ryan~T. McDonald. 2021.
\newblock Zero-shot neural passage retrieval via domain-targeted synthetic
  question generation.
\newblock In \emph{Proceedings of the 16th Conference of the European Chapter
  of the Association for Computational Linguistics: Main Volume, {EACL} 2021,
  Online, April 19 - 23, 2021}, pages 1075--1088.

\bibitem[{Mintz et~al.(2009)Mintz, Bills, Snow, and
  Jurafsky}]{mintz2009distant}
Mike Mintz, Steven Bills, Rion Snow, and Dan Jurafsky. 2009.
\newblock Distant supervision for relation extraction without labeled data.
\newblock In \emph{Proceedings of the Joint Conference of the 47th Annual
  Meeting of the ACL and the 4th International Joint Conference on Natural
  Language Processing of the AFNLP}, pages 1003--1011.

\bibitem[{Nguyen et~al.(2016)Nguyen, Rosenberg, Song, Gao, Tiwary, Majumder,
  and Deng}]{nguyen2016ms}
Tri Nguyen, Mir Rosenberg, Xia Song, Jianfeng Gao, Saurabh Tiwary, Rangan
  Majumder, and Li~Deng. 2016.
\newblock Ms marco: A human generated machine reading comprehension dataset.
\newblock In \emph{CoCo@ NIPS}.

\bibitem[{O{\u{g}}uz et~al.(2021)O{\u{g}}uz, Lakhotia, Gupta, Lewis, Karpukhin,
  Piktus, Chen, Riedel, Yih, Gupta et~al.}]{ouguz2021domain}
Barlas O{\u{g}}uz, Kushal Lakhotia, Anchit Gupta, Patrick Lewis, Vladimir
  Karpukhin, Aleksandra Piktus, Xilun Chen, Sebastian Riedel, Wen-tau Yih,
  Sonal Gupta, et~al. 2021.
\newblock Domain-matched pre-training tasks for dense retrieval.
\newblock \emph{arXiv preprint arXiv:2107.13602}.

\bibitem[{Qu et~al.(2021)Qu, Ding, Liu, Liu, Ren, Zhao, Dong, Wu, and
  Wang}]{qu2021rocketqa}
Yingqi Qu, Yuchen Ding, Jing Liu, Kai Liu, Ruiyang Ren, Wayne~Xin Zhao, Daxiang
  Dong, Hua Wu, and Haifeng Wang. 2021.
\newblock \href {https://doi.org/10.18653/v1/2021.naacl-main.466}
  {{R}ocket{QA}: An optimized training approach to dense passage retrieval for
  open-domain question answering}.
\newblock In \emph{Proceedings of the 2021 Conference of the North American
  Chapter of the Association for Computational Linguistics: Human Language
  Technologies}, pages 5835--5847, Online. Association for Computational
  Linguistics.

\bibitem[{Reddy et~al.(2021)Reddy, Yadav, Sultan, Franz, Castelli, Ji, and
  Sil}]{reddy2021towards}
Revanth~Gangi Reddy, Vikas Yadav, Md~Arafat Sultan, Martin Franz, Vittorio
  Castelli, Heng Ji, and Avirup Sil. 2021.
\newblock Towards robust neural retrieval models with synthetic pre-training.
\newblock \emph{arXiv preprint arXiv:2104.07800}.

\bibitem[{Robertson and Zaragoza(2009)}]{robertson2009probabilistic}
Stephen Robertson and Hugo Zaragoza. 2009.
\newblock \emph{The probabilistic relevance framework: BM25 and beyond}.
\newblock Now Publishers Inc.

\bibitem[{Sachan et~al.(2021)Sachan, Patwary, Shoeybi, Kant, Ping, Hamilton,
  and Catanzaro}]{sachan2021end}
Devendra~Singh Sachan, Mostofa Patwary, Mohammad Shoeybi, Neel Kant, Wei Ping,
  William~L Hamilton, and Bryan Catanzaro. 2021.
\newblock End-to-end training of neural retrievers for open-domain question
  answering.
\newblock \emph{arXiv preprint arXiv:2101.00408}.

\bibitem[{Shinoda et~al.(2021)Shinoda, Sugawara, and Aizawa}]{shinoda2021can}
Kazutoshi Shinoda, Saku Sugawara, and Akiko Aizawa. 2021.
\newblock \href {https://doi.org/10.18653/v1/2021.mrqa-1.6} {Can question
  generation debias question answering models? a case study on
  question{--}context lexical overlap}.
\newblock In \emph{Proceedings of the 3rd Workshop on Machine Reading for
  Question Answering}, pages 63--72, Punta Cana, Dominican Republic.
  Association for Computational Linguistics.

\bibitem[{Tsatsaronis et~al.(2015)Tsatsaronis, Balikas, Malakasiotis, Partalas,
  Zschunke, Alvers, Weissenborn, Krithara, Petridis, Polychronopoulos
  et~al.}]{tsatsaronis2015overview}
George Tsatsaronis, Georgios Balikas, Prodromos Malakasiotis, Ioannis Partalas,
  Matthias Zschunke, Michael~R Alvers, Dirk Weissenborn, Anastasia Krithara,
  Sergios Petridis, Dimitris Polychronopoulos, et~al. 2015.
\newblock An overview of the bioasq large-scale biomedical semantic indexing
  and question answering competition.
\newblock \emph{BMC bioinformatics}, 16(1):1--28.

\bibitem[{Yang et~al.(2018)Yang, Qi, Zhang, Bengio, Cohen, Salakhutdinov, and
  Manning}]{yang2018hotpotqa}
Zhilin Yang, Peng Qi, Saizheng Zhang, Yoshua Bengio, William Cohen, Ruslan
  Salakhutdinov, and Christopher~D. Manning. 2018.
\newblock \href {https://doi.org/10.18653/v1/D18-1259} {{H}otpot{QA}: A dataset
  for diverse, explainable multi-hop question answering}.
\newblock In \emph{Proceedings of the 2018 Conference on Empirical Methods in
  Natural Language Processing}, pages 2369--2380, Brussels, Belgium.
  Association for Computational Linguistics.

\bibitem[{Zhu et~al.(2021)Zhu, Lei, Wang, Zheng, Poria, and
  Chua}]{zhu2021retrieving}
Fengbin Zhu, Wenqiang Lei, Chao Wang, Jianming Zheng, Soujanya Poria, and
  Tat-Seng Chua. 2021.
\newblock Retrieving and reading: A comprehensive survey on open-domain
  question answering.
\newblock \emph{arXiv preprint arXiv:2101.00774}.

\end{thebibliography}
\bibliographystyle{acl_natbib}

\appendix

\section{Parameter Details}
\label{appendix:parameter}
For the pre-training, all models we reproduced are trained with 20 million Q-P pairs. Specifically, our reported \name~is trained on the combination of 10 million DL pairs and 10 million CM pairs while the \name~(DL) and \name~(CM) reported in Table \ref{table:ood} are trained on 10 million DL pairs and 10 million CM pairs, respectively. More parameters details are shown in the table below.

\begin{table}[H]
\centering
\scalebox{0.8}{
\begin{tabular}{ccc}
\hline
Hyperparameter         & Pre-training & Fine-tuning  \\ \hline
Epoch                  & 5            & 40           \\
Batch Size             & 400          & 256          \\
GPU Resource           & 32GB GPU × 8 & 32GB GPU × 8 \\
Learning Rate          & 2e-5         & 2e-5         \\
Warmup Ratio           & 0.1          & 0.1          \\
Learning Rate Decay    & Linear       & Linear       \\
Shared Encoder         & True         & False        \\
Maximum Q Length       & 150          & 256          \\
Maximum P Length       & 256          & 256          \\ \hline
\end{tabular}
}
\end{table}

\section{Negative Sampling}
\label{appendix:negative}
While negative sampling plays an import role in contrast learning, we have explored different types of negatives to pair with queries: 
(1) Random negatives: passages randomly selected from the corpus 
(2) Overlap negatives: passages have entity overlap with queries but fail to match either DL or CM topology. Our experimental results in Table \ref{table:negative} show that the model perform better when it adopts random negatives. We conjecture that the overlap negatives may be too hard for the self-supervised pre-training. Thus, we pair one random negative to each query during pre-training. 

\begin{table}[H]
\centering
\scalebox{0.7}{
\begin{tabular}{ccccccc} 
\hline
\multirow{2}{*}{\begin{tabular}[c]{@{}c@{}}Negative \\Type\end{tabular}} & \multicolumn{3}{c}{NQ}                           & \multicolumn{3}{c}{TriviaQA}                      \\
                                                                         & top5           & top20          & top100         & top5           & top20          & top100          \\ 
\hline
None                                                                     & 45.7           & 66.0           & 79.9           & 62.6           & 75.2           & 83.0            \\
Random                                                                   & \textbf{51.2} & \textbf{70.2} & \textbf{82.0} & \textbf{65.9} & \textbf{76.9} & \textbf{84.0}  \\
Overlap                                                                  & 49.7           & 67.8           & 80.3           & 63.1           & 75.1           & 83.0            \\
\hline
\end{tabular}}
\caption{Top-k zero-shot retrieval accuracy of \name~using different types of negatives during pre-training.}
\label{table:negative}
\end{table}

\section{Data Analysis on NQ Samples}
\label{appendix:analysis_evidence}
We discuss how we conduct data analysis to determine the hyperlink-based topology. Driven by a strong interest in what roles the Q-P overlapping spans play, we conduct exploratory data analysis on the widely-used NQ dataset. Specifically, we extract all entities and mentions from the Q-P pairs using TagMe \cite{ferragina2010tagme} for further investigation. We observe about 55\% queries $q$ either explicitly mentions the titles of $p$ or successfully links to the document via TagMe. This observation motivates us to construct the dual-link topology where the pseudo queries $q$ mention $p$ via a hypertext. Moreover, we observe about 45\% queries $q$ do not mention the titles of $q$ but instead they share the same mentions. This encourages us to adopt the co-mention topology where the pseudo $q$ and $p$ both mention a third-party document through hypertext.

\section{Pseudo Code for \name~Pairs}
\label{appendix:pseudo}
\begin{algorithm}
    \renewcommand{\algorithmicrequire}{\textbf{Input}}
	\renewcommand{\algorithmicensure}{\textbf{Notation:}}
    \SetKwFunction{FMain}{Main}
    \SetKwFunction{FA}{IsDL}
    \SetKwFunction{FB}{IsCM}
	\begin{algorithmic}
	\Ensure \\

$q, p \gets$ Wikipedia passages \\
$t_{Q} \gets$ Topical entity of passage $q$ \\
$\mathcal M(q) \gets$ The set of entities mentioned in $q$ \\
$d_{in}(q) \gets$ in-degree of the Wikipedia entity $t_Q$ \\
$K \gets$ in-degree threshold for CM pairs 
    \end{algorithmic}
    
    \SetKwProg{Fn}{Def}{:}{}
    \Fn{\FA{$q$, $p$}}{
        \If{$t_{P} \in \mathcal M(q) \And t_{Q} \in \mathcal M(p)$}{\Return 1}
        \Else{\Return 0}
    }
\;
    \SetKwProg{Fn}{Def}{:}{}
    \Fn{\FB{$q$, $p$}}{
        \ForEach {$m \in \mathcal M(q)$}{
            \If{$d_{in}(m)<K \And \;
                m \in M(p) \And \;
                t_{Q} \in M(p) $}{\Return 1}
            \Else{\Return 0}
        }
    }
	\caption{\name~Pairs Identification} 
\end{algorithm}

\section{Fact-level Evidence Reduction}
\label{appendix:reduction}
Intuitively, we assume any mentioned entity, let's say $e_{Y}$ mentioned in a Wikipedia document $X$, is used to describe the topical entity $e_{X}$ of this document. In other words, $e_{Y}$ is likely to attend in a topically relevant fact or event related to $e_{X}$, which can be represented as a triple <$e_{X}$,~$r_{XY}$,~$e_{Y}$> where $r_{XY}$ is a latent relation between $e_{X}$ and $e_{Y}$. 

Given any passage pair ($q$, $p$) from Wikipedia, we consider $q$ and $p$ have fact-level evidence if they both entail a fact that can be represented as a triple, let's say <$e_{X}$,~$r_{XY}$,~$e_{Y}$>. Further, if both passages $q$ and $p$ contain representative hypertext or topic of $e_{X}$ and $e_{Y}$, we consider such fact-level evidence can be induced by hyperlink-based topology, namely hyperlink-induced fact. Below we show that any Q-P pair with hyperlink-induced fact while satisfying answer containing is within either DL or CM hyperlink-based topology. 

Following the example above, given $q$ and $p$ containing a factual triple <$e_{X}$,~$r_{XY}$,~$e_{Y}$>, we have facts <$e_{Q}$,~$r_{QX}$,~$e_{X}$>, <$e_{Q}$,~$r_{QY}$,~$e_{Y}$> at $q$-side while <$e_{P}$,~$r_{PX}$,~$e_{X}$>, <$e_{P}$,~$r_{PY}$,~$e_{Y}$> at $p$-side. Further, $p$ entails <$e_{P}$,~$r_{PQ}$,~$e_{Q}$> because of the answer containing property. 

Case1: $e_{P} = e_{X}$ or $e_{P} = e_{Y}$. Then $q$ entails facts <$e_{Q}$,~$r_{QP}$,~$e_{P}$>. Note that $r_{QP}$ is likely but not necessarily to be identical to $r_{PQ}$ in $p$. In this case, ($q$, $p$) fits in the Dual-link topology in our definition.

Case2: $e_{P} \neq e_{X}$ and $e_{P} \neq e_{Y}$. Then given the facts <$e_{Q}$,~$r_{QX}$,~$e_{X}$> at $q$-side, and  <$e_{P}$,~$r_{PX}$,~$e_{X}$> at $p$-side, ($q$, $p$) fits in the Co-mention topology.

\section{Case Studies on Multi-hop Retrieval}
\label{appendix:hotpot}
We evaluate \name~on multi-hop scenario where knowledge from different documents need to be associated. Besides significant improvements shown in Table \ref{table:hotpot}, we conduct case study to investigate its capability on knowledge-intensive retrieval. In Table \ref{table:hotpot-case}, a complex question is proposed,  requiring the retriever firstly to retrieve the document \textit{``Apple Remote''} and then \textit{``Front Row (software)''}. \name~successfully retrieves both golds in the top-10 retrieved passages while the vanilla DPR fails. We find 6 items retrieved by \name~are related to the brand \textit{``Apple''} while 4 by DPR, which indicates stronger comprehension and associative ability of \name.

\begin{table}[H]
\centering
\scalebox{0.7}{
\begin{tabular}[c]{|p{10cm}|}
\hline
\textbf{Question:} \\
Aside from the Apple Remote, what other device can control the program Apple Remote was originally designed to interact with? \\ 
\hline
\textbf{Evidence Passage:}  \\
1. \textcolor{blue}{Apple Remote}: The Apple Remote is a remote control device ... was originally designed to interact with the \textcolor{blue}{Front Row} media ...   \\
2. \textcolor{blue}{Front Row (software)}: Front Row is a discontinued media ... is controlled by an Apple Remote or the \textcolor{red}{keyboard function keys} ... \\ 
\hline
\textbf{Top-10 Retrieved Titles (Ours):} \\
\textcolor{blue}{Apple Remote}; ITunes Remote; Remote computer; Wii Remote; Remote Shell; Kinect; Siri Remote; \textcolor{blue}{Front Row (software)}; Apple TV; Spinning pinwheel;  \\ 
\hline
\textbf{Top-10 Retrieved Titles (DPR's):} \\
\textcolor{blue}{Apple Remote}; ITunes Remote; Console (video game CLI); Control Panel (Windows); Apple Wireless Keyboard; Chooser (Mac OS); Remote computer; Media player (software); ToggleKeys; Button (computing); \\
\hline
\end{tabular}}
\caption{Case studies on HotpotQA dataset. Blue gives the titles of gold passages while red gives answer span.}
\label{table:hotpot-case}
\end{table}

\section{Case Studies on Q-P Paraphrase}
\label{appendix:case}
We present case studies on the constructed \name~Q-P pairs in Table \ref{table:dl-case} and Table \ref{table:cm-case}. 
As we can see, there are entity- and fact-level paraphrasing across questions and passages, which can be interpreted as factual evidence for passage matching in OpenQA. 
For example, entity-level variants such as \textit{``Robert and Richard Sherman''} vs. \textit{``Sherman Brothers''}, and fact-level paraphrases such as \textit{``Abby Kelley and Stephen Symonds Foster ... working for abolitionism''} vs. \textit{``... radical abolitionists, Abby Kelley Foster and her husband Stephen S. Foster''} can be found in our examples.

\begin{table*}[hb]
\centering
\scalebox{0.51}{
\begin{tabular}{|l|l|} 
\hline
\multicolumn{1}{|c|}{\textbf{Query}}
&
\multicolumn{1}{c|}{\textbf{Passage}}\\ 
\hline
\begin{tabular}[c]{@{}l@{}}
Title: \textbf{Abby Kelley}\\ 
Liberty Farm in \textcolor{blue}{Worcester, Massachusetts}, the home of\\
Abby Kelley and Stephen Symonds Foster, was designated\\
a National Historic Landmark because of its association\\
with their lives of working for abolitionism.
\end{tabular}              
&
\begin{tabular}[c]{@{}l@{}}
Title: \textbf{Worcester, Massachusetts}\\ 
Two of the nation's most radical abolitionists,~\textcolor{blue}{Abby Kelley Foster}~and her husband Stephen S. Foster, adopted Worcester as their home,\\
as did Thomas Wentworth Higginson, the editor of The Atlantic Monthly and Emily Dickinson's avuncular correspondent, and Unitarian\\
minister Rev. Edward Everett Hale. The area was already home to Lucy Stone, Eli Thayer, and Samuel May Jr. They were joined in their\\
political activities by networks of related Quaker families such as the Earles and the Chases, whose organizing efforts were crucial to ...
\end{tabular}\\ 
\hline
\begin{tabular}[c]{@{}l@{}}
Title: \textbf{Callisto Corporation}\\ 
They were best known for their series of computer games\\
for the Macintosh in the 1990s, including \textcolor{blue}{ClockWerx}, Spin\\
Doctor, Super Maze Wars and Super Mines.
\end{tabular} 
&
\begin{tabular}[c]{@{}l@{}}
Title: \textbf{ClockWerx}\\
ClockWerx is a computer game created by~\textcolor{blue}{Callisto Corporation}~that was released in 1995. The game was originally released by Callisto\\
under the name SSpin Doctor". Later, with some game play enhancements, it was published by Spectrum HoloByte as "Clockwerx\\
which was endorsed by Alexey Pajitnov according to the manual. A 3DO Interactive Multiplayer version was planned but never released.\\
The object of the game is to solve a series of increasingly difficult levels by swinging a rotating wand from dot to dot until the player reaches\\
the "goal" dot. Enemy wands ...
\end{tabular}\\ 
\hline
\begin{tabular}[c]{@{}l@{}}
Title: \textbf{Sivaji Ganesan}\\
Some of his famous hits during this period are " \textcolor{blue}{Vasantha}\\
\textcolor{blue}{Maligai~} ", " Gauravam ", " Thanga Pathakkam " and\\
" Sathyam ".
\end{tabular}
&
\begin{tabular}[c]{@{}l@{}}
Title: \textbf{Vasantha Maligai}\\
Vasantha Maligai is a 1972 Indian Tamil -language romance film, directed by K. S. Prakash Rao and produced by D. Ramanaidu . The film\\
stars \textcolor{blue}{Sivaji Ganesan}~and Vanisri , and is the Tamil remake of the 1971 Telugu film " Prema Nagar ". " Vasantha Maligai " was released on\\
29 September 1972 and became a major commercial success, running in theatres for nearly 750 days. A digitally restored version of the film\\
was released on 8 March 2013, and another one on ...
\end{tabular}\\
\hline
\begin{tabular}[c]{@{}l@{}}
Title: \textbf{Say Anything (band)}\\
Around this time, the band also released " \textcolor{blue}{Alive with the}\\
\textcolor{blue}{Glory of Love} "~as a single.
\end{tabular}
&
\begin{tabular}[c]{@{}l@{}}
Title: \textbf{Alive with the Glory of Love}\\
"Alive with the Glory of Love" is the first single from~\textcolor{blue}{Say Anything}~\textbackslash{}'s second album " ...Is a Real Boy ". "Alive with the Glory of Love"\\
was released to radio on June 20, 2006. The song was a hit for the band, charting at number twenty-eight on the Alternative Songs chart.\\
The song, described as an "intense and oddly uplifting rocker about a relationship torn by the Holocaust," by the " Pittsburgh Post-Gazette ",\\
is actually semi-biographical in nature, telling the story of songwriter and vocalist Max Bemis \textbackslash{}'s ...
\end{tabular}\\
\hline
\begin{tabular}[c]{@{}l@{}}
Title: \textbf{Dorothy Sue Hill}\\
Hill taught home economics from 1960 to 1969 for the\\
\textcolor{blue}{Allen Parish School } \textcolor{blue}{Board}~and from 1969 to 1992 for the\\
Beauregard Parish School Board .
\end{tabular}
&
\begin{tabular}[c]{@{}l@{}}
Title: \textbf{Allen Parish School Board}\\
Allen Parish School Board is a school district headquartered in Oberlin in Allen Parish in southwestern Louisiana , United States. From\\
1960 to 1969,~\textcolor{blue}{Dorothy Sue Hill}, the state representative for Allen, Beauregard , and Calcasieu parishes, taught home economics for Allen\\
Parish schools.
\end{tabular}\\
\hline
\end{tabular}
}
\caption{Examples of DL Q-P pairs where text in blue gives evidence and answer.} 
\label{table:dl-case}
\end{table*}

\begin{table*}[hb]
\centering
\scalebox{0.51}{
\begin{tabular}{|l|l|} 
\hline
\multicolumn{1}{|c|}{\textbf{Query}}
&
\multicolumn{1}{c|}{\textbf{Passage}}                                                          \\ 
\hline
\begin{tabular}[c]{@{}l@{}}
Title: \textbf{Daniel Gormally}\\
In 2015 he tied for the second place with \textcolor{blue}{David Howell}\\
and Nicholas Pert in the 102nd British Championship \\andeventually finished fourth on tiebreak.\end{tabular}
&
\begin{tabular}[c]{@{}l@{}}
Title: \textbf{Nicholas Pert}\\
In 2015, Pert tied for 2nd–4th with \textcolor{blue}{David Howell} and \textcolor{red}{Daniel Gormally}, finishing third on tiebreak, in the British Chess Championship and\\
later that year, he finished runner-up in the inaugural British Knockout Championship, which was held alongside the~London Chess Classic.\\
In this latter event, Pert, who replaced Nigel Short after his late~withdrawal, eliminated~Jonathan Hawkins in the quarterfinals and Luke\\
McShane in the semifinals, then he lost to David Howell 4–6 in the final.\end{tabular}  \\ 
\hline
\begin{tabular}[c]{@{}l@{}}
Title: \textbf{Ojuelegba, Lagos}\\
Ojuelegba is a suburb in~\textcolor{blue}{Surulere~}local government area\\
of Lagos State.\end{tabular}
&
\begin{tabular}[c]{@{}l@{}}
Title: \textbf{Simi (singer)}\\
... on September 8, 2017. Her third studio album " Omo Charlie Champagne, Vol. 1 " was released to coincide with her thirty-first birthday\\
on April 19, 2019. She launched her record label Studio Brat in June 2019. Simi was born on 19 April 1988 in \textcolor{red}{Ojuelegba}, a suburb of\\
\textcolor{blue}{Surulere}, Lagos State, as the last of four children. In an interview with Juliet Ebirim of " Vanguard " newspaper, Simi revealed that her\\
parents separated when she was 9 years old. She also revealed that she grew up as a ...\end{tabular} \\ 
\hline
\begin{tabular}[c]{@{}l@{}}
Title: \textbf{The Aristocats}\\
Longtime Disney collaborators \textcolor{blue}{Robert and Richard}\\
\textcolor{blue}{Sherman} composed multiple songs for the film, though\\
only two made it in the finished product.
\end{tabular}
&
\begin{tabular}[c]{@{}l@{}}
Title: \textbf{In Search of the Castaways}\\
Later sang the \textcolor{blue}{Sherman Brothers}~\textbackslash{}' theme song \textbackslash{}' \textcolor{red}{The Aristocats}~\textbackslash{}' from Disney\textbackslash{}'s 1970 animated film "The Aristocats". Ïn Search of the\\
Castaways" was a commercial success. Upon its initial release, it earned \$4.9 million in North American theatrical rentals. It was one of the\\
12 most popular movies at the British box office in 1963. " The New York Times " declared: Ït is, as we say, a whopping fable, more\\
gimmicky than imaginative, but it doesn\textbackslash{}'t lack for lively melodrama that is more innocent and wholesome than much of the ...
\end{tabular}\\ 
\hline
\begin{tabular}[c]{@{}l@{}}
Title:~\textbf{Jang Jin-young}\\
As of 2008, Jang was one of the highest paid stars in the\\
\textcolor{blue}{Korean film industry}, earning in the region of per film.
\end{tabular}
&
\begin{tabular}[c]{@{}l@{}}
Title:~\textbf{Scent of Love}\\
Scent of Love (Scent of Chrysanthemums) is a 2003 \textcolor{blue}{South Korean film}, and the directorial debut of Lee Jeong-wook. The film is based on\\
a novel of the same name by Kim Ha-in, and stars~\textcolor{red}{Jang Jin-young}~and Park Hae-il in the lead roles. Like her character, Jang Jin-young\\
battled stomach cancer and died in 2009. The film received an around of 900,000 admissions nationwide and on May 16, 2003 the film\\
was screened at the Cannes Film Festival. University student Seo In-ha meets a ...
\end{tabular}\\ 
\hline
\begin{tabular}[c]{@{}l@{}}
Title:~\textbf{Vera Menchik}\\
Vera Menchik ("Vera Frantsevna Menchik" 16 February\\
1906 – 26 June 1944) was a Russian-born British-\\
Czechoslovak chess player who became the first\\
\textcolor{blue}{women\textbackslash{}'s world chess champion}.
\end{tabular}
&
\begin{tabular}[c]{@{}l@{}}
Title:~\textbf{Paula Wolf-Kalmar}\\
Paula Wolf-Kalmar (11 April 1880 - 29 September 1931) was an Austrian chess master, born in Zagreb. She took 5th at Meran 1924\\
(unofficial European women's championship won by Helene Cotton and Edith Holloway). After the tournament three of the participants\\
(Holloway, Cotton and Agnes Stevenson) defeated three others (Kalmar, Gulich and Pohlner) in a double-round London vs. Vienna match.\\
She was thrice a~\textcolor{blue}{Women's World Championship}~Challenger. She took 3rd, behind \textcolor{red}{Vera Menchik} and Katarina Beskow at London 1927 ...
\end{tabular}\\
\hline
\end{tabular}}
\caption{Examples of CM Q-P pairs where text in blue gives evidence while red gives answer.} 
\label{table:cm-case}
\end{table*}

\end{document}